\documentclass[twocolumn,11pt]{article}
\usepackage[utf8]{inputenc}

\usepackage[letterpaper,top=2.25cm,bottom=2.25cm,left=2cm,right=2cm,marginparwidth=1.75cm]{geometry}

\usepackage{comment}

\usepackage{mathtools}
\usepackage{amssymb}
\usepackage{siunitx}
\usepackage{textcomp}

\usepackage{bbding}
\usepackage{authblk}
\usepackage{svg}
\usepackage{graphicx}
\usepackage[export]{adjustbox}
\usepackage{stfloats}
\usepackage{fbox}
\definecolor{reviewblue}{RGB}{0,50,130}
\definecolor{citecolor}{RGB}{90,90,100}
\usepackage[colorlinks=true,linkcolor=black,urlcolor=black,citecolor=citecolor]{hyperref}

\usepackage[labelfont=bf, labelsep=period]{caption}

\DeclareCaptionFormat{myformat}{#1#2#3\hrulefill}
\newenvironment{arial}{\fontfamily{ppl}\selectfont}{\par}

\usepackage{natbib}

\definecolor{reviewblue}{RGB}{0,50,180}

\title{\vspace{0cm}\huge \textbf{Dynamic neural manifolds for flexible closed-loop control on neuromorphic hardware}}

\author[1,2]{Oskar von Seeler}
\author[1,2]{Christian Tetzlaff}
\author[1,2,3, \Envelope]{Andrew B. Lehr}

\affil[1]{\small Group of Computational Synaptic Physiology, Department of Neuro- and Sensory Physiology,
University Medical Center G{\"o}ttingen, Germany}
\affil[2]{\small Campus Institute Data Science,
University of G{\"o}ttingen, Germany}
\affil[3]{\small Circulant Labs,
Bensheim, Germany}
\affil[\Envelope]{Correspondence: andrew@circulant.ai}

\date{\vspace{-0.9cm}}

\begin{document}\raggedbottom
\maketitle

\begin{abstract}
\begin{arial}
    \noindent \textbf{In biological circuits, sequential neural activity evolves along dynamic, low-dimensional manifolds to enable flexible behavior. 
    Spiking network models link aspects of this sequential activity to features of manifold geometry through specific circuit mechanisms, making dynamic neural manifolds parameterizable, and thereby offering an explainable framework for neural computation.
    Extending this framework to neuromorphic engineering, we present an implementation on the SpiNNaker 2 chip for real-time, closed-loop control. 
    By allowing sensory inputs to modulate heterogeneous inhibition, gain, and transient currents, our architecture drives rapid subspace rotations to switch between behaviors, as well as fine-grained trajectory control within them. 
    We validate this via a robotic simulation where an agent uses sensory feedback to dynamically reconfigure its manifold geometry to navigate through a maze. 
    Our results establish dynamic manifolds as a feasible approach for explainable neuromorphic architectures and a substrate for investigating biological neural dynamics.
    }
\end{arial}
\end{abstract}

\section*{Introduction}

Rapid improvements in parallel, brain-inspired hardware offer an opportunity to achieve the energy efficiency and adaptability of biological organisms in artificial systems.
Realizing this potential requires an accompanying suite of architectures and algorithms that draw from the computational logic of the brain.
Central to this vision is the development of robust design frameworks that can synthesize complex circuit components into functional, explainable behavior.

The language of neural manifolds provides a formal geometric framework that may support this synthesis in neuromorphic engineering.
In this view, the collective activity of a population of $N$ neurons is represented as a trajectory in an $N$-dimensional state space, where each neuron defines a single dimension. 
Activity in the brain is typically constrained to low-dimensional manifolds that capture the latent variables of the task \citep{gao2017theory, jazayeri2021interpreting, langdon2023unifying}.
The geometric features of these manifolds map to behavioral execution, for example with subspace rotations taking place when switching between movements or behaviors \citep{sabatini2024reach, elsayed2016reorganization, tang2020minimally} or the speed of neural trajectories adapting with changes in movement timing \citep{wang2018flexible, rodriguez2024motor}.
By mapping how circuit-level mechanisms generate a particular geometry, we can design autonomous systems where the internal state is mathematically interpretable, supporting explainable artificial intelligence.

\begin{figure*}[!ht]
    \centering
    \includegraphics[width=0.8\linewidth]{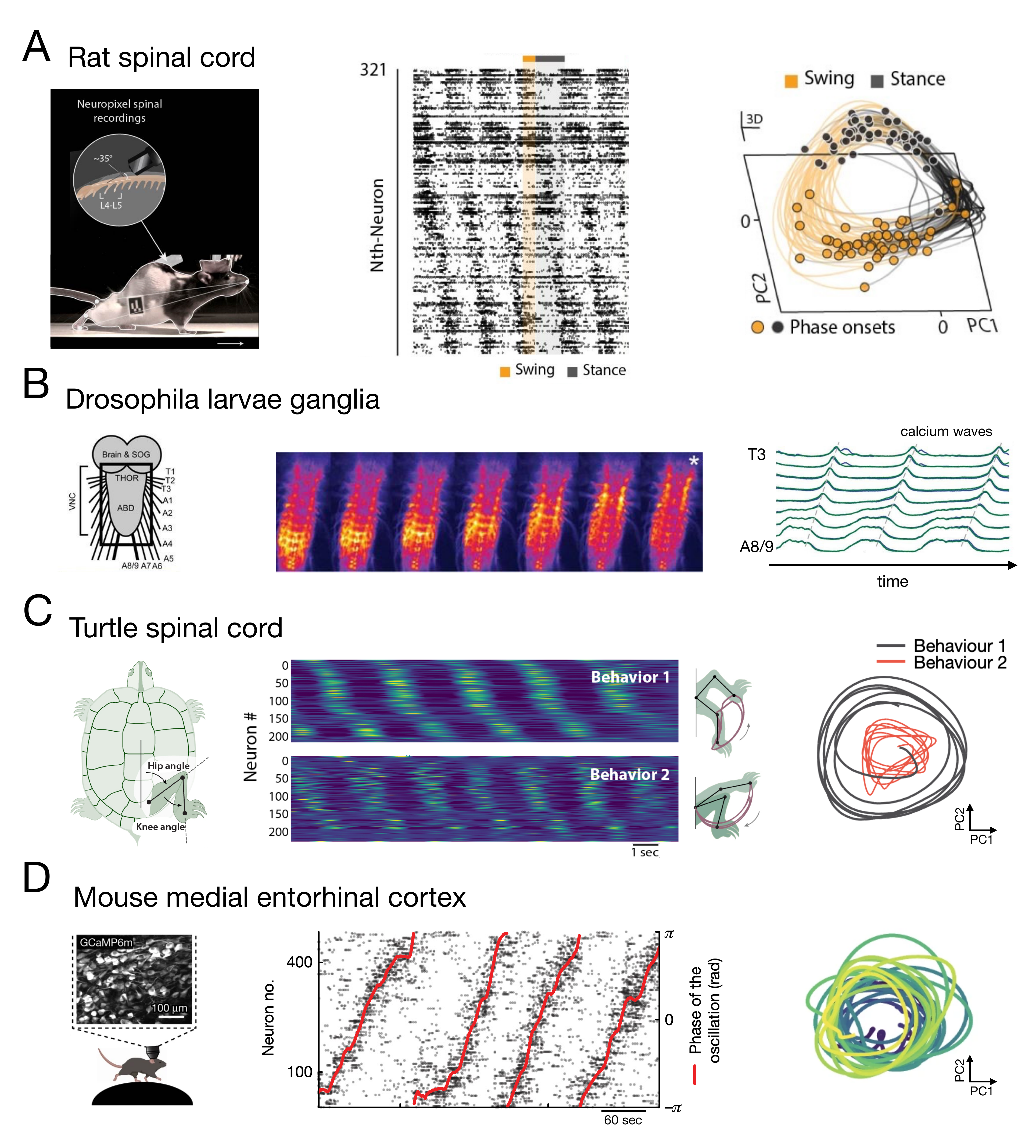}
    \caption{Oscillatory sequences on low dimensional manifolds across the brain and spinal cord. 
    Left panels show schematic of experimental setting, middle panels show sequences, right panels show cyclic behavior in state space (A,C,D) or as traces (B).
    \textbf{A} Rat spinal cord.
    \textbf{B} Drosophila larvae ganglia. 
    \textbf{C} Turtle spinal cord.
    \textbf{D} Mouse medial entorhinal cortex.
    Reproduced under CC BY license from \citep{komi2026neural,pulver2015imaging,linden2021movement,gonzalo2024minute}.}
    \label{fig:sequences}
\end{figure*}

At the same time, what we observe directly in the brain is not a manifold, but patterns of neural activity. 
Across diverse neural structures and species, ranging from the low-level motor control circuits of the spinal cord \citep{linden2022movement, komi2026neural} up to higher-order cognitive structures like the medial entorhinal cortex (MEC) \citep{gonzalo2024minute}, a common observation is neural activity which is structured as dynamic but reliable sequences (Figure \ref{fig:sequences}).
These sequences are not static, but rather, they undergo dynamic changes in response to internal states and external demands. 
By modulating the temporal structure of neural sequences, biological circuits flexibly reorganize their activity to accommodate a wide repertoire of behaviors.

Because a neural manifold is the geometric manifestation of a population’s progression through a sequence, the parameters of that sequence directly define the geometry and dynamics of the manifold. 
We have recently developed a framework  \citep{lehr2024dynamic,lehr2025spatially} identifying specific circuit mechanisms that function as ``control knobs" for these geometric properties. 
In particular, heterogeneous inhibition facilitates subspace reorientation, allowing a sequence to rotate into new hyperplanes to switch behavioral states. 
Simultaneously, gain modulation and transient synaptic currents regulate sequence propagation, providing direct control over the velocity and curvature of the neural trajectory. 
The map between low-level circuit architecture and high-level geometric features offers a set of design principles for engineering autonomous systems with predictable, explainable behaviors.

\begin{figure*}[!hb]
    \centering
    \includegraphics[width=0.9\linewidth]{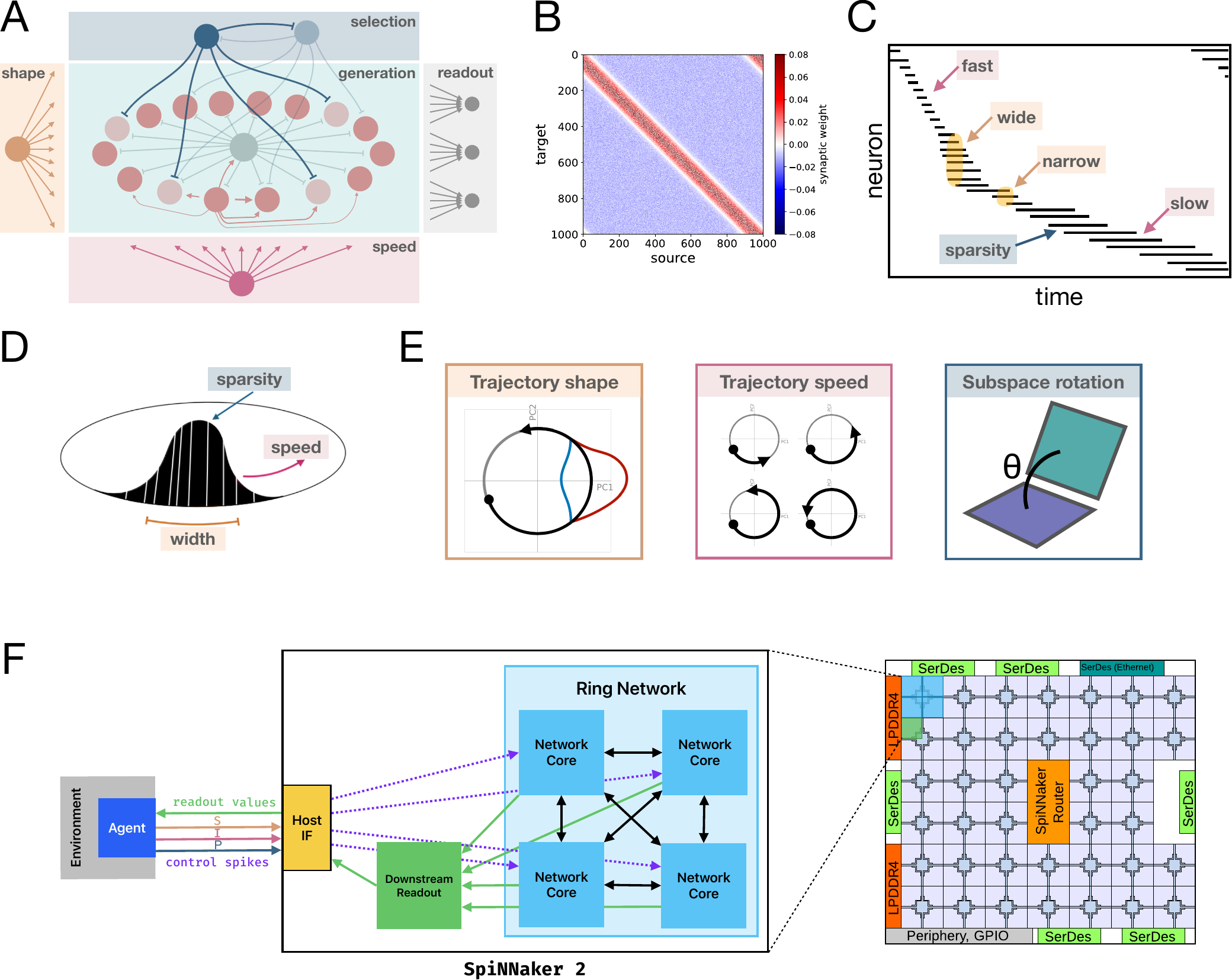}
    \caption{Implementation overview.
    \textbf{A} Network architecture. Control neurons (speed, shape, selection) change the activity within a ring network, which can be decoded by downstream readout neurons.
    \textbf{B} Circulant weight matrix with 50\% connection sparsity (weights scaled $2\times$ to compensate).
    \textbf{C} Sketch of typical activity pattern under varying inputs from control neurons. Control neurons determine the speed of propagation around the ring, the width of the activity bump and the activation sparsity.
    \textbf{D} Schematic of the activity bump traveling around the ring, characterized by properties controlled by the control neurons.
    \textbf{E} Effect of control inputs on low-dimensional manifold of neural activity.
    \textbf{F} SpiNNaker 2 hardware implementation. Ring network and readout are implemented on-chip and communicate through the host interface with an external system to interact with the agent in the environment (SpiNNaker 2 diagram reproduced from~\cite{gonzalez2024spinnaker2}).}
    \label{fig:implementation}
\end{figure*}

Here we apply this design framework and demonstrate its utility through a closed-loop implementation on the SpiNNaker 2 neuromorphic chip~\cite{mayr2019spinnaker, hoppner_spinnaker_2022}. 
In a proof-of-principle application, a two-wheeled robotic agent navigates a maze, utilizing local environmental cues as feedback to the dynamic manifold circuit. 
With sensory inputs modulating circuit mechanisms like inhibition and gain in real-time, the agent can dynamically reconfigure its manifold geometry to switch between behavioral states (e.g., steering vs. jumping) and adjust its velocity and trajectory during execution of the task.
Our work offers an energy-efficient, low-latency substrate for adaptive neuromorphic control while simultaneously providing a computational testbed for exploring how biological circuits translate spatiotemporal neural dynamics into goal-directed, embodied behavior.

\section*{Results}

Here we set out to implement a circuit architecture for the dynamic control of neural manifolds on neuromorphic hardware (Figure \ref{fig:implementation}). 
The goal is an explainable parameterization of neural activity and manifold features as well as their closed-loop control by sensory input, which can in turn be mapped onto computational primitives required for a task.
We focus on the concrete example of moving through a maze, requiring that an agent navigates obstacles and turns along the way, a typical test of autonomous artificial agents and world model architectures \citep{Beattie2016DeepMindL,pasukonis2022evaluating,hafner2023mastering}. 
In this work we assume that the agent has already learned a world model and focus on translating this world model into motor control via neuroscience-inspired dynamic neural manifolds.

For this, we consider a canonical activity pattern observed across the spinal cord and brain --- oscillatory sequences of neural activity on the timescale of seconds or longer (Figure \ref{fig:sequences}). 
For coordinated movement, sequences of this nature arise in the spinal cord and regions like the motor cortex and are thought to coordinate motor neuron activation to generate the appropriate sequence of muscle contractions.

These sequences can be modeled as a bump of neural activity moving along a ring of neurons (Figure \ref{fig:implementation}), a canonical object in computational neuroscience.
Recent work has shown that a set of circuit mechanisms acting on the ring (Figure \ref{fig:implementation}A) can dynamically control aspects of the sequential activity (Figure \ref{fig:implementation}C,D), which smoothly map onto predictable changes at the level of the neural manifold (Figure \ref{fig:implementation}e; \citep{lehr2024dynamic,lehr2025spatially}).
In the N-dimensional state space of neural activity, these dynamic control mechanisms allow the circuit to push the state into regions of state space that are easily separable by a downstream readout as well as smoothly adjust trajectories in this space to control movement parameters, ultimately allowing dynamic and adaptable behavior.

In this work we first map this circuit architecture to the neuromorphic chip SpiNNaker 2 and refine the implementation for efficiency (Figure \ref{fig:implementation}F). 
We make use of SpiNNaker 2's ability to stream data to generate a closed loop, enabling sensory information to dynamically control the circuit via the dynamic control mechanisms, shaping the neural manifold as required by task demands.

\subsection*{Implementation of dynamic neural manifolds on SpiNNaker 2}

Our architecture for dynamic neural manifolds is implemented as a ring network with asymmetric recurrent connectivity, which leads to the formation of a stable bump of activity that progresses along the ring, resulting in oscillatory sequences of neural activity (Figure \ref{fig:implementation}).
We further implement neural mechanisms that control the shape and speed of the ongoing neural trajectory as well as the orientation of the low dimensional subspace within the ambient space (Figure \ref{fig:implementation}A,C-E), based on~\cite{lehr2024dynamic,lehr2025spatially}, with some modifications to the model to achieve better runtime performance when executed on the SpiNNaker 2 system.

The original model is formulated with rate-based neurons. However, as the SpiNNaker 2 hardware is optimized for spike-based communication~\cite{hoppner_spinnaker_2022}, we added a spike-based communication layer between the neurons in the ring and implemented a probabilistic conversion of rates to spikes for each neuron, treating the rate as the probability of the neuron to spike in the current time step.
This increases performance by significantly reducing communication within the distributed system.
To further improve performance, we introduced sparsity into the connectivity matrix to reduce the number of incoming spikes to be processed at each neuron (Figure \ref{fig:implementation}B).

To make efficient use of the limited on-chip memory, we utilize the circulant structure of the weight matrix to avoid storing duplicate weights and instead store a single row of the weight matrix and a sparsity mask indicating with one bit whether a synapse exists or not.
However, the limited local memory still remains a constraint to the implementation.
When storing any information over time, such as control parameter sequences or recordings of spikes or internal rates within the limited SRAM memory, the maximum possible execution time is significantly limited.
As an example, recording the internal rates of 32 neurons on one core would fill up the entire \qty{128}{\kilo\byte} of SRAM within 1000 time steps or one second of simulation time at typical \qty{1}{\milli\second} time steps.
Instead, our implementation allows control parameters to be streamed into the chip during execution while streaming spikes from the SpiNNaker 2 chip to a host system.
Thus actions can be driven by output spikes, which in turn result in sensory inputs that dynamically adapt control parameters to drive subsequent action.

\subsection*{Dynamically controlling neural trajectories and manifolds on chip}

\begin{figure*}
    \centering
    \includegraphics[width=0.8\linewidth]{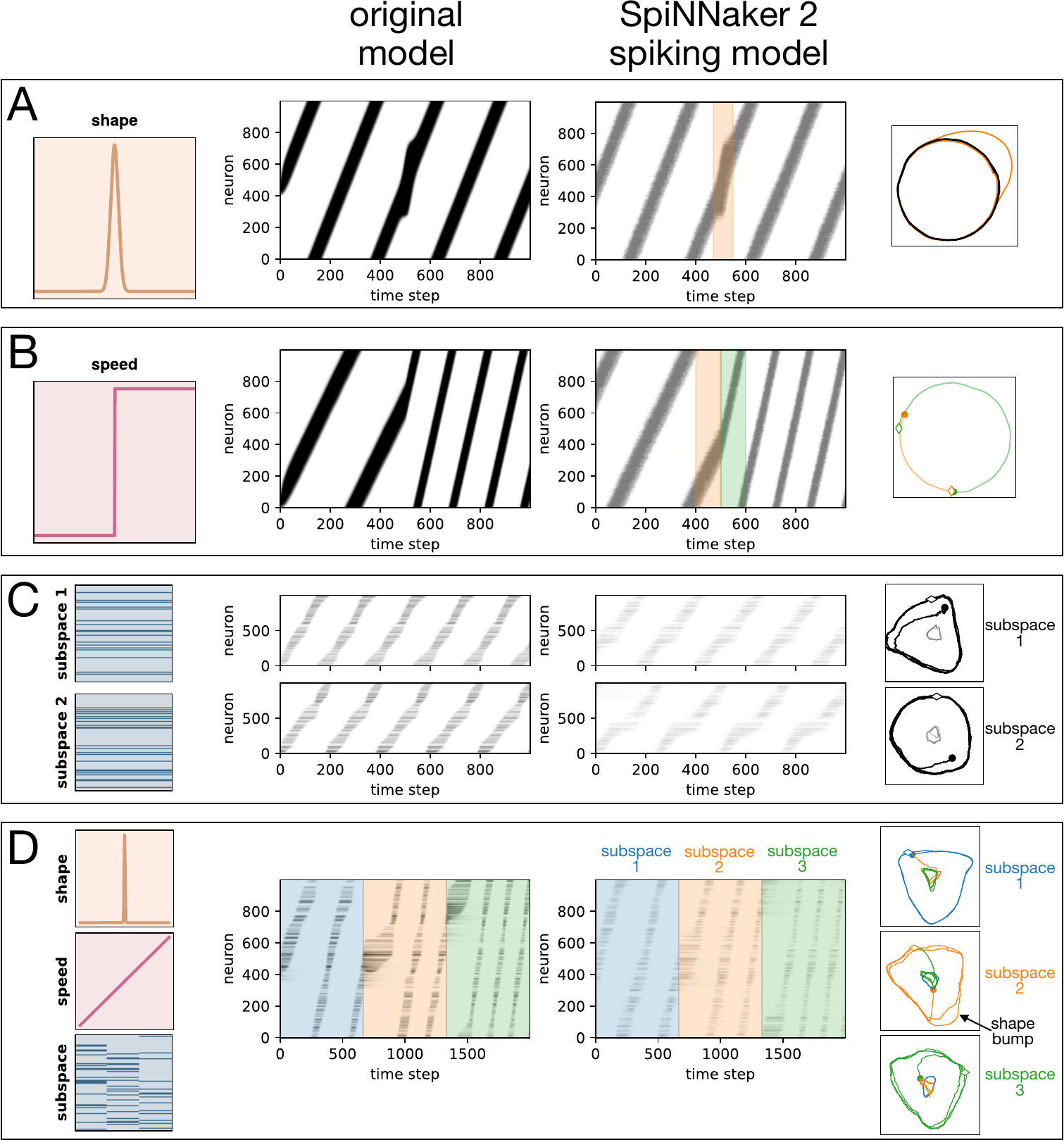}
    \caption{
    Reproducing control sequences from~\cite{lehr2024dynamic} on SpiNNaker 2. For our implementation on SpiNNaker 2 we show the spikes that are used in neuron-to-neuron communication and PCA projections, which are computed on the recorded internal rates. Network sizes follow the original figures, we use 50\% connection sparsity in all cases.
    \textbf{A} Shape control. Additive current $I$ is temporarily increased, causing a change in bump width, which is visible both in recorded spikes and the PCA projection.
    \textbf{B} Speed control. Multiplicative gain $S$ is increased from a low level to a higher level after half of the simulation duration. At higher speed, projection into PC-space reveals larger travel distance of the activity around the ring within the same time frame.
    \textbf{C} Subspace inhibition. Two different subspaces with 80\% selective inhibition each. 
    Sequence progression is undisturbed by inhibition.
    Projecting neural activity of one sequence into the PC-space of another results in a small projection (gray trajectories) compared to the within subspace trajectories (black), demonstrating that both temporal dynamics and behavioral state are robustly represented.
    \textbf{D} All controls combined. Three different subspaces at 80\% subspace inhibition are used, each being active for one third of the total simulation. Speed control $S$ is increased throughout the simulation, resulting in the number of rotations around the ring increasing for each subspace. One temporary increase in the additive current $I$ is given, resulting in an increase in bump size during subspace 2, visible in PC space.
    }
    \label{fig:dynamic-control}
\end{figure*}

\begin{figure*}
    \centering
    \includegraphics[width=\linewidth]{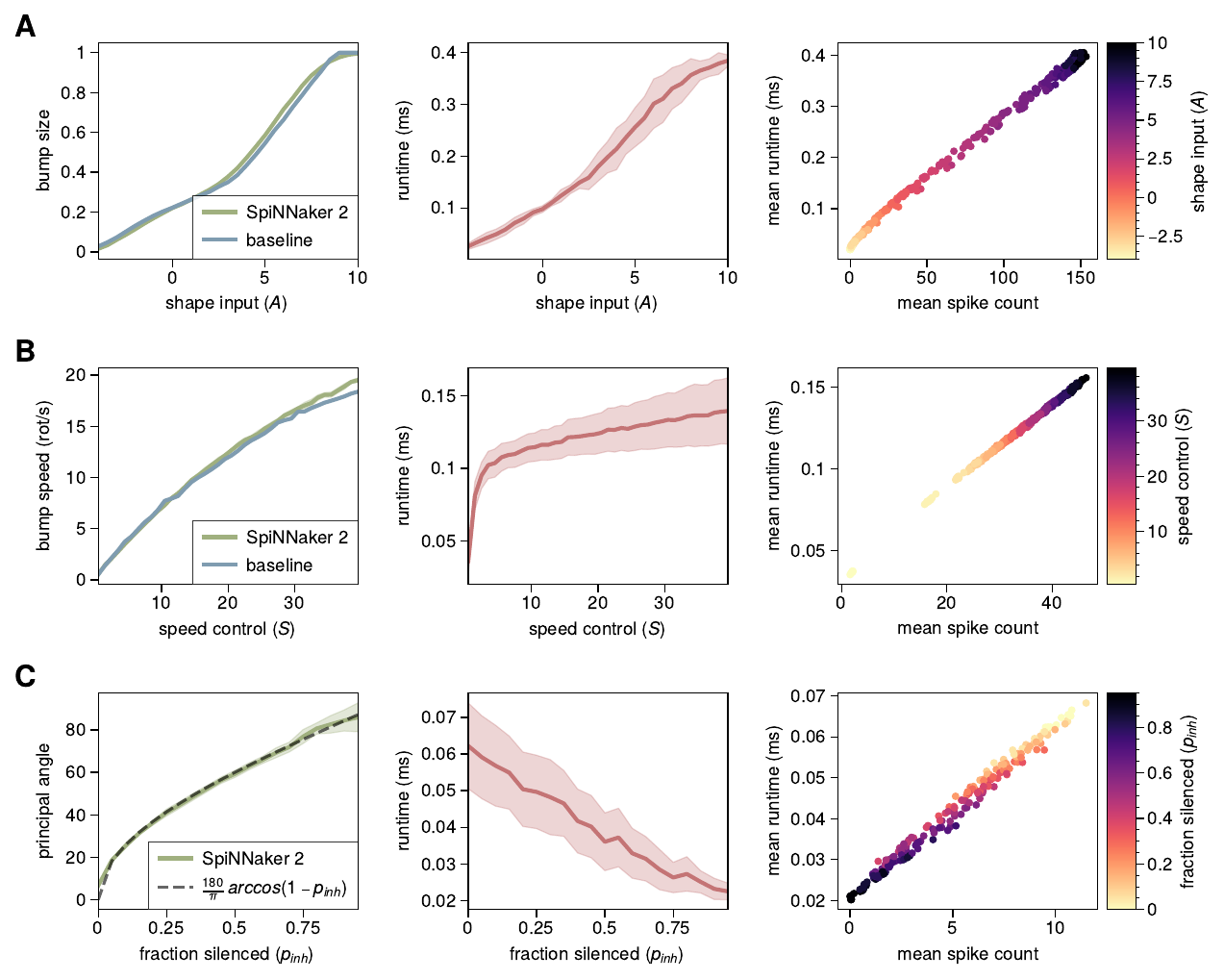}
    \caption{Comparison of the original model and the SpiNNaker 2 implementation across ranges of three control parameter values. All experiments use 500 neurons with 20\% connectivity and are run for 1000 time steps. Each combination of parameters is tested for 10 trials, with each trial using a different random initialization. For the three control parameters we report adherence to the baseline model (first column) and the runtime performance on SpiNNaker 2 depending on that control parameter (second column). The third column compares spike count to the runtime performance.
    \textbf{A} Shape Control $I$. We look at the change in bump size induced by a temporary change in the shape control parameter $I$ (compare Figure \ref{fig:dynamic-control}A). In our spiking implementation, we use the number of spikes as an estimate for the bump size, while we use the sum of rates in the original baseline model.
    \textbf{B} Speed Control $S$. For the speed control we present a constant control input to the network and measure the resulting speed of the bump as full rotations around the ring over time.
    \textbf{C} Fraction of silenced neurons $p_\text{inh}$. Instead of comparing to a baseline model, we compute the first principal angle between two different subspaces with the same inhibition ratio and compare the result to the analytical solution~\cite{lehr2024dynamic}.
    }
    \label{fig:control-parameters}
\end{figure*}

Next we validate our implementation across a range of parameter values and compare to simulations based on the original model \citep{lehr2024dynamic, lehr2025spatially}.
The model exposes three distinct parameters (multiplicative gain, additive current, random silencing) that can control the neural activity sequence and, by this, the neural trajectory and manifold (Figure \ref{fig:implementation}A).
We conducted a set of simulations with our spike-based and sparsified SpiNNaker 2 implementation in which we varied each of these parameters and compared the effects to simulations with the original rate-based, fully-connected model on CPU (Figure \ref{fig:dynamic-control}; Figure \ref{fig:control-parameters}).

\textit{Trajectory shape}: Additive current $I$ can increase or decrease the total number of active neurons, changing the spatial extent or size of the bump traveling around the ring. This implies a control of the radius of the neural trajectory (Figure \ref{fig:dynamic-control}A).
To test the response to this parameter, we provide a gaussian-shaped excitatory or inhibitory additive current $I(t)$, with amplitude $A$, which causes a temporary change in the size of the bump. When $A>0$, the bump size and trajectory radius are increased (Figure \ref{fig:dynamic-control}A).
We estimate bump size by the number of spikes in the SpiNNaker 2 implementation, and as the sum of all non-zero rates in the original rate-based model.
When comparing the models over a range of different values for $A$, the SpiNNaker 2 implementation yields similar responses to changes in this parameter when compared to the original model (Figure \ref{fig:control-parameters}A), confirming the feasibility of dynamically controlling trajectory shape on SpiNNaker 2.

\textit{Trajectory speed}: Dynamic changes in multiplicative gain $S$ affects how quickly the bump travels around the ring, which maps to different neural trajectory speeds (Figure \ref{fig:dynamic-control}B).
This parameter amplifies inputs as gain increases, resulting in faster responses and thus faster progression of the sequence \citep{lehr2024dynamic}. 
For the systematic analysis of the speed control parameter $S$, we provide a constant value for the full simulation and report the number of rotations of the bump within one second. We observe matching behavior between the two models until $S>30$, where activity was marginally faster on SpiNNaker 2 (Figure \ref{fig:control-parameters}B).
Thus the speed of the sequence and resulting neural trajectory in our SpiNNaker 2 implementation can be reliably controlled by modulations in multiplicative gain.

\textit{Trajectory rotation}: The network is equipped with a set of inhibitory ensembles, each of which strongly inhibits a random subset of neurons, silencing them completely (Figure \ref{fig:control-parameters}C).
Inhibition is uniformly random and parameterized by a single parameter, the fraction of neurons silenced $p_{inh} \in [0,1]$.
Switching between inhibitory ensembles changes the subset of neurons supporting the sequence while maintaining the sequential dynamics \citep{lehr2024dynamic,lehr2025spatially}.
This dynamic switching rotates the neural subspace in which most of the dynamics of the neural activity (the manifold) happens, which in turn enables downstream readouts to differentiate each orientation of the manifold.
This enables implementing each behavioral readout as neural trajectories traversing each unique orientation of the manifold.
The angle between these orientations depends on the fraction of neurons silenced by the inhibitory ensembles and grows as $\arccos{1-p_{inh}}$ \citep{lehr2024dynamic}.

To test whether our implementation reproduces this scaling behavior, we varied the fraction of neurons silenced by each inhibitory ensemble and computed the first principal angle between pairs of subspaces.
In particular, we applied PCA to the sequential activity under silencing by each of the 10 inhibitory ensembles, retained the first k=2 components in each case, and computed the first principal angle between each pair of subspaces.
This analysis shows that our SpiNNaker 2 implementation precisely follows the expected scaling behavior in agreement with the rate-based CPU implementation (Figure \ref{fig:control-parameters}C).
Thus, subspace rotations via selective inhibition work robustly and predictably in our SpiNNaker 2 implementation. 

\textit{General Trajectory Manipulation}: All the controls discussed before can be combined to generate complex trajectories required for supporting flexible behavior.
Figure \ref{fig:dynamic-control}D shows one example of varying speed $S$ and shape $I$ concurrently, while switching between three different subspaces.

\subsection*{Efficiency of implementation as a function of the control parameters}

With the optimizations made for deployment on chip, we are able to simulate networks well within the typical \qty{1}{\milli\second} time step duration of applications on SpiNNaker 2.
The runtime scales with parameters of the network, in particular the total number of neurons, connection sparsity, and the number of simultaneously active neurons (number of spikes).
To assess this scaling behavior, we ran a set of simulations for different network sizes and connectivity (Supplementary Figure 1) as well as variations in the parameter space (speed, shape, silencing) (Figure~\ref{fig:control-parameters}).

For the shape-input experiment, the duration per time step increases with shape control parameter $A$ and the corresponding bump size, demonstrating that larger bumps require more computational overhead (Figure \ref{fig:control-parameters}A).
Multiplicative gain $S$ indirectly affects the bump size, increasing marginally with $S$ and in turn increasing processing time (Figure \ref{fig:control-parameters}B). 
Conversely, increasing the subspace inhibition parameter $p_\text{inh}$ silences more neurons, reducing the overall spike count and lowering the runtime (Figure~\ref{fig:control-parameters}C).
Across all three control parameters, the execution time scales linearly with the number of spikes processed within a single time step (Figure~\ref{fig:control-parameters}, third column). 
For the network size and connectivity assessed here (500 neurons, 20\% connectivity, $p_\text{inh} \in [0,0.95]$, $S \in [1,2]$, and $I \in [0, 2]$) our SpiNNaker 2 implementation consistently operates well below the \qty{1}{\milli\second} real-time threshold.
Increasing network size or total number of connections shifts this runtime upward, yet runtime remains low across a range of these parameters (Supplementary Figure 1).

\subsection*{Navigating a virtual maze via closed loop manifold control on chip}

With a working implementation of the core circuit architecture for dynamic neural manifolds, next we aimed to integrate it into an application with closed loop control.
In particular, we consider an agent traversing a virtual maze.
The agent has two independently controlled wheels, enabling forward movement, curved forward movement and turning in place.
Further, the agent is able to jump over obstacles.
The processing of sensory inputs, long-term planning and selection of movements is handled by an external program, which then modifies the control parameter to the ring network.
Motor control is learned directly from the activity patterns within the ring network.

For the ring network in this application, we use 500 neurons, recurrently connected with 20\% connection probability. We construct three subspaces, each with 40\% (200 neurons) of the total 500 neurons being active.
Each subspace encodes for one specific action, moving forward, turning in place or jumping.
Speed and shape control encode movement speed and direction within these distinct actions.

\begin{figure*}
    \centering
    \includegraphics[width=0.95\linewidth]{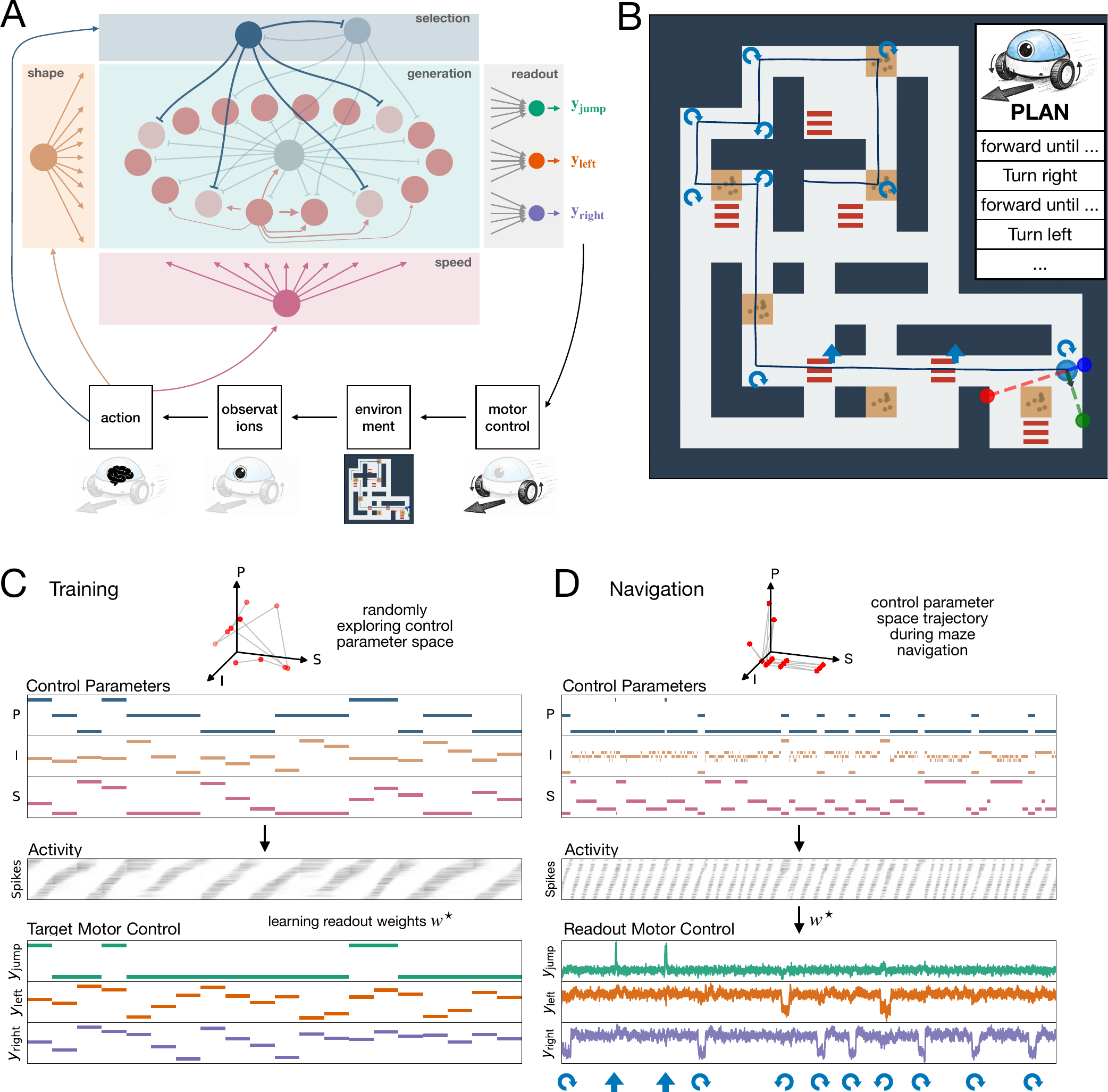}
    \caption{Dynamic neural manifolds for closed-loop control in a virtual maze. An external agent program observes the current state and the action plan and selects control parameters for the ring network. Given these parameters and its internal state, the ring network produces output spikes.
    \textbf{A} Architecture of the closed loop system. A linear readout directly reads motor control values from the ring network activity. These motor control values affect the state of the agent in the environment. New observations are incorporated into the planning by the external policy program, which produces a new set of control parameters.
\textbf{B} Example of a maze with the path an agent took through it. Jumps and turns are highlighted.    
    \textbf{C} Training procedure for the readout weight. A sequence of random actions is selected, and translated into control parameters and target motor controls. Each action is performed for several time steps. The readout weights are fitted to predict the target motor controls from the recorded network activity during the training period. 
    \textbf{D} The activity of the control inputs $P$, $S$, and $I$, the ring network, readouts, and resulting motor controls during maze navigation at test time. Jumps and and turns are highlighted in the motor control sequence, where a turn requires the motors to turn in opposite directions.
    }
    \label{fig:closed-loop-agent}
\end{figure*}

During training, the readout weights are learned through random exploration of the control parameter space (Figure~\ref{fig:closed-loop-agent}C).  200 actions are randomly selected, with each action specifying target motor controls. The actions are mapped to control parameters, and each action is performed for \qty{250}{\milli\second}.
The readout weights are then trained to map from the resulting ring network activity (spikes) to the target motor controls.

During navigation of the maze, the agent must complete steps of a high-level plan. This plan includes steps like \textit{move west until reaching a wall} (Figure~\ref{fig:closed-loop-agent}B).
In addition to this plan, local adaptions are made based on sensory inputs, which include distances to walls in front of and to the sides of the agent, as well as the type of ground (e.g. floor, hurdle, or gravel).
The local adaptations keep the agent on track and avoid crashes.
Combining the current step in the plan with local adaptions, a movement is selected.
This movement is then translated into network control parameters for the circuit mechanisms that control the neural activity trajectory.
At the motor readout, the previously learned readout weights are used to directly inform the motors based on the ring network activities (Figure~\ref{fig:closed-loop-agent}D).
With that setup, the agent is able to integrate the high-level plan into a manifold representation suitable for downstream readout, allowing the agent to traverse the maze successfully (Figure~\ref{fig:closed-loop-agent}B).

Notably, while at first glance the spiking activity during navigation may look like a repeating sequence (Figure~\ref{fig:closed-loop-agent}D), instead there is a hidden code inside the temporal dynamics, made clear when observed from the geometric perspective, we have outlined in previous sections.
The activity is a flexible combination of robust temporal dynamics and sensory/behavioral variables compressed into one conjunctive representation.

\section*{Discussion}

For neuromorphic computing and engineering, a set of compositional building blocks is required to enable plug-and-play development and interoperability of solutions, both in academic and industry settings.
Even more, computational frameworks that define \textit{how} these building blocks should behave are highly critical for coherent development and sustainable progress.
Here we provide a proof of principle for such a building block capable of dynamic motor control based on neuroscientific properties.
In particular, we suggest that our dynamic neural manifold framework may provide a strategy to effectively design explainable and interoperable neural algorithms and implementations on neuromorphic hardware systems.

The predictable activity pattern generated by the ring network and its geometric interpretation on a low-dimensional manifold both contribute to explainability of the algorithm and the explainability of the system as a whole using this in their output control.
Simple neural mechanisms --- gain, inhibition, transient input currents --- readily implementable on neuromorphic hardware enable the dynamic control of parameters of the neural activity sequence. 
Changes in the speed of the sequence, the number of co-active neurons in the moving bump of activity, as well as the subset of neurons receiving selective inhibition steer the neural trajectory through the ambient neural state space in a fully predictable way specified by our dynamic neural manifold framework.

Here we show that using these neural mechanisms as ``control knobs", that can be influenced by sensory inputs or feedback, produces a robust but ever-changing sequence of activity that is fully interpretable in neural state space.
These changes in sequence activity can then be read out to direct motor control during, for instance, a navigation task.
Note that sequences of this nature have been observed in experimental settings \citep[e.g.][]{linden2022movement, gonzalo2024minute, pulver2015imaging, komi2026neural}, reorientation of the neural manifold and modulation of trajectories have been linked to behavioral variables \citep[e.g.][]{sabatini2024reach, rodriguez2024motor}, and theoretical work has addressed mechanisms for implementing these phenomena in a biologically inspired and neuromorphic compatible way \citep{lehr2024dynamic, lehr2025spatially}.

Compared to the relatively simple movement of the two-wheeled robot in our example, the sequential activity patterns will expose their full potential within more complex motion systems such as animals. 
Here the dynamic activity generated by the ring network, akin to the patterns of sequential neural activity observed in the context of motor control across species (Figure \ref{fig:sequences}), might enable downstream readouts to learn the required coordination of complex sequences of muscle contractions and relaxations.
For neuromorphic computing, our approach may find utility in designing efficient and explainable neuro-inspired control of complex architectures with many degrees of freedom, like humanoid and bio-inspired robots.

Here we assumed that the world model and thus the plan for navigating the maze was already learned and focused our efforts on executing the plan, including steering, jumping, and turning, given local sensory information.
Future work will aim to learn a world model on top of the motor control layer.
Particularly interesting for this step may be neuro-inspired architectures that learn from sequences of observations \citep{george2021clone,raju2024space}.
Even more, the brain depends on sensorimotor loops involving multiple subnetworks and networks may communicate via particular subspaces. 
Our dynamic manifold framework extends naturally to interfacing multiple networks ``speaking the same language."

While the ring network is powerful, networks in the brain of animals live in 3D space.
For example cortical layers make approximately 2D sheets with dendrites extending into a third dimension, forming local recurrent connections that are asymmetrical but correlated \citep{ye2026brain}.
Locally correlated asymmetric connections lead to waves of neural activity traveling across the cortex \citep{spreizer2019from, ye2026brain}.
In this setting, the network has more degrees of freedom to generate sequential patterns and due to the heterogeneity in 2D space, some regions of the network are more susceptible than others to modulation, allowing dynamic control of circuit logic \citep{wernecke2026controlling}.
Further mechanisms, like cooperation and competition between sequences \citep{stober2025competition} and their interpretation in neural state space may further increase the repertoire of building blocks in a growing framework for sequence arithmetic.

On the technical side, future extensions could also improve efficiency and autonomy. 
By either utilizing the larger DRAM for long-term recordings, or implementing processes for control and readout directly on the chip to remove the need for recording data.
It could further be possible to utilize the structure of the ring to provide a more efficient implementation, enabling control of speed, shape and subspace without directly simulating each neuron of the ring with its connections.
For now, sensing, higher-level planning, network control and motor control are executed on a separate machine communicating with the ring network SpiNNaker 2 system, however these tasks could be implemented on the same chip for a more integrated and efficient agent control system. 

Taken together, here we applied a framework for neuro-inspired dynamic manifold control \citep{lehr2024dynamic,lehr2025spatially} for a closed-loop control problem on neuromophic hardware. 
We find that the circuit architecture and mechanisms required for dynamically steering the manifold are readily implementable on the SpiNNaker2 chip.
While this work lays a first building block, frameworks for manifold control and manifold steering \citep{lehr2024dynamic,lehr2025spatially,wurgaft2026manifold} in intelligent systems offer promise for explainability, robustness, and efficiency in neuromorphic closed-loop control.

\section*{Methods}
\subsection*{SpiNNaker 2 implementation}

Distributed simulations of neuronal networks on SpiNNaker 2 require assigning groups of neurons from the network to individual processing elements (PEs).
How many neurons are put on one PE is limited by the available local memory at each PE to store neuron parameters, state, connectivity and other data.
Another limitation is the fixed processing time per time step (typically \qty{1}{\milli\second}), which means processing all inputs and updating all neurons on one PE should not take longer than that time period.

In the case of the  ring network we assign 32 neurons to one PE.
Each of these neurons has an internal state given by a rate $r_i \in [0,1]$. This rate is updated according to
\begin{align}
    x_i^{(t+1)} &= \sum_j {W_{ji} r_j^{(t)}} + I^{(t)} \\
    r_i^{(t+1)} &= F\left(r_i^{(t)} + \frac 1\tau \left(-r_i^{(t)} + p_i^{(t)} S^{(t)} x_i^{(t+1)}\right) \right)
\end{align}
\noindent where $W_{ji}$ is the weight of the synapse connecting neuron $i$ to neuron $j$ and $I$ is an external input provided to every neuron.
$x_i$ collects all synaptic inputs to the neuron, which are then scaled by the speed control $S$, $p_i$ indicating whether neuron $i$ is in the active, non-inhibited subset and the time constant $\tau$.
$F$ clamps the values of $r$ to the range of $[0,1]$.

To make better use of the messaging infrastructure of SpiNNaker 2 we utilized spike-based communication between neurons instead of directly computing with the rates of other neurons.
The modified model still computes a rate as an internal measure for the activity of a neuron.
Binary spikes are generated based on the rates, which are used for the neuron-to-neuron communication.
To generate spikes based on the rate of a neuron, neurons spike stochastically at time step $t$ with probability $r^{(t)}/2$. In this model a rate $r=1$ corresponds to a spike frequency of \qty{500}{\hertz}.
Instead of actually randomly deciding for each neuron whether it spikes or not, which would require expensive random number generations, we devised a deterministic rule that approximately matches the spike frequency.
Neuron $i$ spikes at time step $t$ if:
\begin{align}
    s_i^{(t)} &= 1 \Leftrightarrow t+i \equiv 0 \pmod{\left\lceil \frac{2}{r_i^{(t)}} \right\rceil}. \\
    \tilde{x}_i^{(t+1)} &= \sum_j {2 W_{ji} s_j^{(t)}} + I^{(t)}
\end{align}

To adjust for the reduced input through spikes that arrive at most every other time step, synaptic weights are doubled.

In addition to spike-based communication, sparse connections further reduce the computational load of running the ring network by reducing the time it takes to process a single incoming spike on a core. 
Two neurons $i$ and $j$ are connected through a directed synapse with probability $p$. Synaptic weights are multiplied by $1/p$ to adjust for the reduced communication caused by the introduced sparsity.

The synapses are randomly generated before running the network and sent to chip as bitmasks.
A core, simulating up to 32 neurons of the network, has one 32 bit bitmask for each neuron in the full network.
A 1 in that bitmask indicates that the pre-synaptic neuron is connected to the neuron at the specific position on the core.
Because the dense weight matrix is circulant, storing this binary connectivity combined with a single row of the weight matrix is sufficient for representing the connectivity within the network:
If two neurons are connected, the weight can be retrieved from the weight row based on the difference of the indices of the two connected neurons: $W_{ji} = w_{j-i \pmod N}$.

Weights for the one row of the weight matrix are stored as signed 8-bit integers, and then scaled by a shared weight exponent.
The weight exponent is derived from the value of largest absolute value in the original weight matrix and depends on several parameters, such as the scaling required because of sparsity and spike frequency.

To run a simulation on the chip several parameters have to be provided.
This includes the bitmasks and weights for the synapses discussed above, as well as initial rates for all neurons. Following~\cite{lehr2024dynamic}
, the initial rates are generated by running a network with the same parameters for a few timesteps, to start with an appropriate bump width and rate distribution.
To allow dynamically switching between subspaces by disabling subsets of neurons, each neuron gets a 32 bit bitmask, indicating whether that neuron is part of a specific subspace.
This allows representing 32 different projections by selecting a different index within that bitmask. 

To control the dynamic activity of the network, three control parameters can be provided for each time step. Trajectory shape control $I$ and speed control $S$ are arbitrary floating point values, and one integer $p \in \{0,1,\dots,31\}$ selects the active projection. Alternatively, these control parameters can be set in a closed loop during execution (see below).

The rate of each neuron can be recorded as a floating point number for each time step and read from chip after execution for analysis.
While the maximum number of neurons to be simulated on one core is 32, with fewer neurons on one core, more memory is available for recording.
With 16 neurons on each core, a network of 2000 neurons and 50\% connectivity can be simulated and recorded for 1000 time steps, or \qty{1}{\second}.

To allow running for longer durations while still recording activity data, spikes can be streamed from the chip to a host computer during runtime.
For that purpose the SpiNNaker 2 chip provides the HostIF interface.
When properly configured, it can be treated like any other processing element with regards to sending spikes to and receiving spikes from it.
Spikes received by the HostIF are forwarded to the host computer via an UDP connection.
Spikes sent in the opposite direction to the SpiNNaker 2 chip are forwarded to the on-chip router, allowing the host to send spikes to any core.
As an example of bandwidth, in our implementation we were able to receive up to $\sim$270 spikes from the chip per \qty{1}{\milli\second} time step.
However, with more advanced buffering methods, we expect a higher throughput to be possible.

To avoid modifying our main ring network implementation and to allow detailed control, such as attaching the current iteration number as metadata, we employ one additional processing element to collect spikes from the network and forward them to the HostIF interface.
This spike streamer introduces one additional timestep of delay when passing on spikes to the host.

As discussed before, control parameters can be updated from the host during runtime, avoiding the need for a control sequence to be stored on the chip and enabling simulations of arbitrary length.
This is implemented by sending a spike with a specific payload structure containing a selector for the updated field and the new value to each core running the dynamic control model from the host:

\begin{itemize}
    \item speed control $S$: signed fixed-point value, sign, 5 integer bits, 10 fractional bits
    \item shape control $I$: signed fixed-point value, sign, 5 integer bits, 10 fractional bits
    \item projection $p$: integer in $\{0,1,\dots,31\}$ selecting one of the predefined projections.
\end{itemize}

Assuming the control program on the host can keep up, the latency of a full control loop can be as low as a few time steps.
If neurons spike at time step $t_0$, the spike streamer forwards it to the HostIF interface at $t_0+1$, and the host receives the spike at $t_0+2$.
Assuming the host can instantly compute new control parameters, these can reach the neurons at time step $t_0+3$.

Without on-chip recording and pre-specified control parameters the memory usage of the neuron model is independent of simulation time, allowing simulations of arbitrary runtime. 

\subsection*{Closed-loop applications}

\subsubsection*{Environment}

The environment consists of a maze, with an agent that has the goal of moving from a starting position to a target. The maze structure is generated using the \textit{labmaze} library~\cite{Beattie2016DeepMindL}, which generates mazes consisting of multiple rooms connected by corridors.
We replace generated spawn points by hurdles and objects by gravel patches.

The agent is a two-wheeled robot, with the speed of each wheel being an independently controlled real number.
While the maze is generated from square tiles on an integer grid, the agent moves in real-valued steps and can move to any position not blocked by a wall.
Further, the robot is able to jump, which is required to pass hurdles in the maze.
The agent can observe distances to the nearest hurdle or wall in three directions: forward, to the left and to the right.
Further, the agent can observe the ground type below its center (floor, hurdle or gravel) and its global heading (angle).

To simulate a real-world scenario with imprecise control, we add noise into the motor control.
Both selected motor speeds are scaled by noise factors $k_\text{left}, k_\text{right} \sim \mathcal{N}(1, \sigma)$, where $\sigma=0.01$ on normal ground and $\sigma=0.1$ on gravel.

The simulation is updated at \qty{1000}{\hertz}, with new observations given to the agent, and movement applied based on the motor speeds selected by the agent.
Speed updates are instantaneous, there is no smoothing on the motor control across timesteps.

\subsubsection*{Agent implementation}
As the focus of this work was to demonstrate the ability of learning motor controls from activations in the ring network and dynamic closed-loop control utilizing the network, the agent does not perform complex  learning procedures to find its way through the maze.
Instead, there is a predetermined, manually created plan for each maze.
This plan includes steps like \textit{``Turn West"} and \textit{``Move forward until reaching the wall"}.
Each step in the plan contains an action and a condition for when to move on the next step.

Outside of this predetermined plan, the agent performs local adaptations in the loop, such as keeping straight and slowing down over gravel to avoid diverging from the path due to the higher movement noise.
Keeping straight involves maintaining the targeted heading and staying clear of the walls to the side. Both goals can be achieved by performing a curved motion by setting slightly different speeds to the two wheels while moving forward.
Another local adaptation the agent performs is adjusting its speed based on the distance to the next wall forward. When coming close to a wall while moving forward, the speed is reduced.

Based on this combination of a global plan and local adaption, the agent policy selects a target action to be performed. This action is encoded into the ring network control parameters, mapping an action to subspace, speed and shape control parameters.
Three different subspaces are used for encoding three different movements: moving forward, turning in place and jumping over a hurdle.
Within each of these movements we linearly map the forward speed to the speed input $S$ of the network and the difference between the two motors to the shape input $I$, with upper and lower bounds on the two inputs to remain in sustainable ranges (such that activity does not die or saturate).

\subsubsection*{Ring Network Details}

For the closed loop application we use a ring network with 500 neurons.
With 32 neurons mapped to one core, this occupies 16 out of the 152 cores in a single-chip system.
This leaves the majority of a single-chip system available for other tasks, such as higher-level planning, sensor processing or an on-chip readout from the ring network.
The neurons are connected with 20\% connectivity, and each subspace has 40\% active neurons.

\subsubsection*{Readout}

Each neuron in the ring network produces a sequence of spike signals $s_i^{(t)}$ (1 iff neuron $i$ spikes at timestep $t$).
Downstream neurons can learn target signals as linear combinations of those spike sequences.

For our readout, the binary spikes are smoothed over time with an exponential moving average for further processing: 
\begin{align}
    \overline{s_i^{(t)}} &= \sum_{k=0}^\infty \alpha^k (1-\alpha) s_i^{(t-k)} \\
    &= (1-\alpha) s_i^{(t)} + \alpha \overline{s_i^{(t-1)}}.
\end{align}
From these smoothed spike signals, motor speeds $y_\text{left}$ and $y_\text{right}$ and whether or not to jump is determined using a linear readout:
\begin{equation}
y_x = \left(
\overline{s_1^{(t)}}, \cdots, \overline{s_n^{(t)}}, 1
\right) \cdot w_x
\end{equation}
\noindent with weight vectors $w_\text{left}$, $w_\text{right}$ and $w_\text{jump}$. For the motor speeds the result is used directly, and the robot tries to jump if $y_\text{jump} > 0.5$.
Equivalently, with potentially reduced computational cost, the individual readout values $y_x$ could be computed directly from the spikes, and then smoothed after the linear readout.

The weight vectors are trained without the agent in the loop: 200 random action with random target motor speeds are sampled. These actions are mapped to ring network control parameters and constantly passed into the network for  250 timesteps for each action.
Spiking activity from the ring during these episodes are recorded while applying exponential moving average smoothing.
Then, the readout weights are trained with linear regression to reconstruct the target readouts from the smoothed activations $\overline{s_i^{(t)}}$.

\section*{Acknowledgements}

\section*{Competing interests}

ABL is a co-founder and shareholder of Circulant GmbH. The remaining authors declare no competing interests.

\bibliographystyle{abbrv}
\bibliography{biblio.bib}

\end{document}


\raggedbottom
\linenumbers
\maketitle

\begin{figure*}[!hb]
    \centering
    \includegraphics[width=0.9\linewidth]{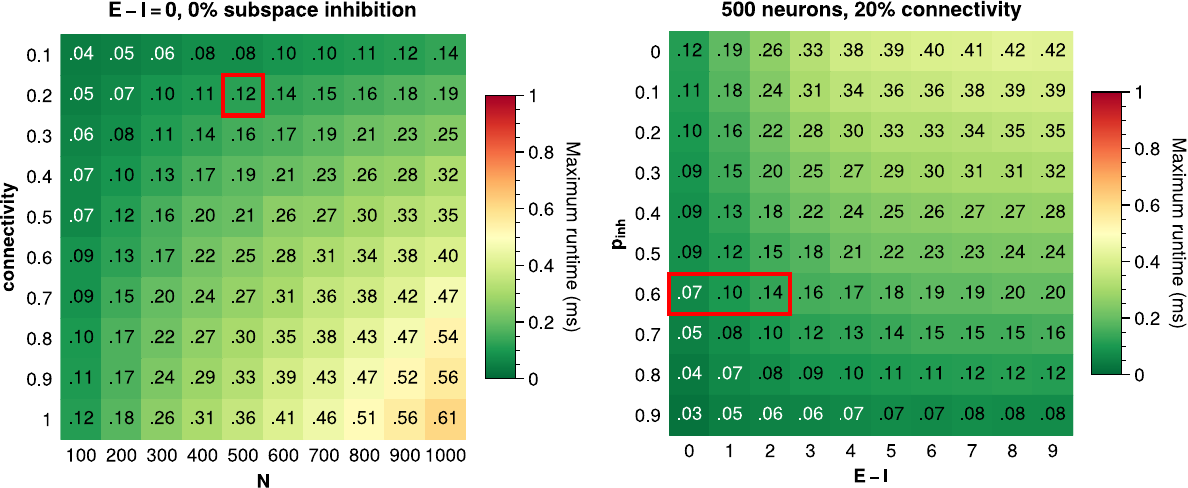}
    \caption{Additional Benchmarks of the ring network implementation on SpiNNaker 2. Red boxes highlight the control parameter ranges used in the closed loop application.
    \textbf{A} Varying the number of neurons and the connectivity within the network. Both a larger number of neurons and a higher connectivity lead to an increase in runtime per timestep on the SpiNNaker 2 system.
    \textbf{B} Varying additive current $I$ and inhibition strength $p_\text{inh}$ simultaneously.
    }
    \label{fig:supp-bench}
\end{figure*}